# What pronouns for Pepper? A critical review of gender/ing in research

**Katie Seaborn**

*Industrial Engineering and Economics*
*Tokyo Institute of Technology*

**Alexa Frank**

*Independent Scholar*

# What Pronouns for Pepper?
# A Critical Review of Gender/ing in Research

Katie Seaborn
*Tokyo Institute of Technology, seaborn.k.aa@m.titech.ca.jp*

Alexa Frank
*Independent Scholar, a.frank93@yahoo.com*

Gender/ing guides how we view ourselves, the world around us, and each other—including non-humans. Critical voices have raised the alarm about stereotyped gendering in the design of socially embodied artificial agents like voice assistants, conversational agents, and robots. Yet, little is known about how this plays out in research and to what extent. As a first step, we critically reviewed the case of Pepper, a gender-ambiguous humanoid robot. We conducted a systematic review (n=75) involving meta-synthesis and content analysis, examining how participants and researchers gendered Pepper through stated and unstated signifiers and pronoun usage. We found that ascriptions of Pepper's gender were inconsistent, limited, and at times discordant, with little evidence of conscious gendering and some indication of researcher influence on participant gendering. We offer six challenges driving the state of affairs and a practical framework coupled with a critical checklist for centering gender in research on artificial agents.

## 1    INTRODUCTION

Within HCI and adjacent spaces, there has been a growing recognition that we must consider socially-constructed and situated characteristics such as gender [4, 40, 111, 116], race [5, 97, 135], and their intersections [128] in design and research practice. This does not just mean the platform being designed or studied and its users: it also includes the creators, researchers, and other stakeholders within the larger communities of practice. In other words, we need to turn the lens on ourselves, as well. This call is not new, especially within HCI spaces. A decade ago, Bardzell wove together feminist theory and interaction design practice under the heading of feminist HCI [4, 28]. She proposed two broad ways of how feminist theory can enhance how we do HCI: through critique and through action. Yet, a recent 10-year review indicated a lack of genuine follow-through within HCI [28]. The notion of *feminism through design* (FtD) was summarily proposed, calling on researchers, practitioners, and educators to explicitly carry out feminist praxis in HCI research [27]. Now that the tides have shifted again and new calls are being made, we have a renewed impetus to do better.

One area in which a self-critical perspective may be necessary is the study of socially embodied artificial agents. Such agents, ranging from voice assistants like Apple's Siri to humanoid robots like Aldebaran-SoftBank's Pepper, are often designed with humanlike features. Of these, gender and the act of *gendering*, or the association of gender with a person, place, or thing, has been recognized as eminent. Decades of research has shown that people tend to react to even subtle gender cues in stereotyped ways, often without realizing it [32, 34, 40, 54, 58, 93, 95, 111, 115, 118, 119, 127, 129, 134, 141, 145]. Humanoid robots, in particular, are designed to look and feel human in form factor, clothing, voice, gestures, and so on. Yet, gender/ing is often overlooked in their design and study [96, 115, 116, 118, 127]. Gender tends to be arbitrarily prescribed by roboticists, who end up perpetuating the status quo [115]. Moreover, people tend to attribute gender even when robots are intended to be gender neutral [34, 134]. On the research side, little is known about how the academic community approaches gender/ing outside of the studies in which it is the subject. Yet, we are surely not exempt from the human tendency to gender our world, even if we are not aware of it. The implications are grave. Without acknowledging gender/ing in research, we cannot know how and to what extent it has influenced previous scholarship and is currently steering the course. As with race in "colour-evasiveness" [1, 97], we cannot be "gender-evasive" and disregard the issue. To do so is risky and raises ethical questions about representation, inclusion, and reinforcing known negative states of affair [95, 96, 116, 150]. Finally, without confronting the issue of gender/ing in our midst head-on, we cannot create a gender-sensitive path forward in the vital and evolving field of human-agent interaction.

This work is a first step towards a more gender-conscious way of doing and thinking in research on socially embodied intelligent agents. The novelty of this work is its self-critical frame: looking at ourselves and our own practice, as well as the experience of participants. Our goal was to uncover how gender/ing has been treated within research in the recent past—after sounding calls could reasonably have been answered. To this end, we chose to anchor our work on Pepper (2014-21), a commercial humanoid robot created by Aldebaran-SoftBank Robotics. We chose Pepper due to anecdotal evidence of ambiguity in ascriptions of its gender. For instance, while its creators' stated aims were for a gender neutral robot [103, 134], official text materials[1] clearly gender Pepper as masculine, i.e., through the use of he/his pronouns. Yet, others have chosen to *not* use Pepper based on their interpretation of its gender as feminine (cf. [63]). Anecdotally, we have encountered different takes on Pepper's gender in conversations with roboticists, technical experts, and the public. We started to ask, more seriously: What gender *is* Pepper—how is it being gendered by people, and is there consensus? Followed closely by: Are we on the research side aware of this phenomenon? Then, to the present work: What is the nature of this phenomenon, and how widespread is it within our own circles? With these questions in mind, we conducted a systematic review of the literature on the social construction of Pepper's gender. We discuss our results with respect to the state of gender sensitivity in HCI and beyond. While we used Pepper as a case study, at a high level, this work is relevant to research on any socially embodied artificial agent. Our main contributions are:

- Empirical evidence of unconscious gendering and diverse ascriptions of gender to Pepper, derived from a corpus of patterns in participant *and* researcher (i.e., academics, designers, practitioners, roboticists, etc.) data;
- Initial evidence that researcher gendering influences participant gendering of Pepper;
- A framework and critical checklist grounded in these findings and informed by critical theories and human-centered design methods to guide theory and practice in the design and study of socially embodied agents.

---

[1] https://www.softbankrobotics.com/emea/en/pepper

We now turn to how gender/ing has been theorized across disciplines of study as well as within the context of HCI, and then particularly with respect to robotic agents like Pepper.

## 2 RELATED WORK

### 2.1 Conceptualizing Gender

Gender is a fundamental aspect of humanity: a universal facet of modern and historical cultures, and present as far back as records allow us to see. But what is gender? The notion of gender is often positioned against the notion of sex. While used analogously in the past [68], these terms have been reimagined within gender studies, feminist theory, sociology, anthropology, and psychology as based in a distinction between biology (*sex*) and culture (*gender*) [45, 68, 114]. Sex refers to the spectrum of visible and invisible physical markers in anatomy, typically associated with reproduction and puberty. People can be male, female, or intersex, having ambiguous characteristics [41]. Gender refers to the social and psychological side as situated within a given cultural context: how people identify (*internally or personally*), express themselves on the outside, e.g., through mannerisms, clothes, etc. (*presentation or expression*), and are perceived by others (*social perception*). Typically, this is within a binary model of masculine and feminine. At a societal level, expectations about a person's behavior, role in society, value, and so on are socially constructed in relation to gender [45] in ways that are dynamic, contested, policed, assumed, weaponized, restricted, subverted, and performed [19, 116]. Most people are assigned a gender at birth in line with their visible sex characteristics, and then raised in accordance with the dominant cultural views of that gender. Those whose gender and sex align with societal expectations are considered cisgender, while those who do not feel that their gender identity matches their apparent sex may identify as transgender, non-binary, genderqueer, or gender fluid [46, 65, 114]. There are also recognized third genders, Two-Spirit [64], and agender people. Notably, some have argued that in practice it can be difficult to separate the biological from the cultural, leading to arguments for a framing of "gender/sex" [55]. Misgendering is a social phenomenon that can occur when someone misreads another person's gender [2]. This can happen purposely or not, for many reasons, but is based in a person's mental models and perceptions of outward gender cues and expressions that are themselves based on social gender norms. In sum, both sex and gender are continuums, and arguably socially constituted as well as socially situated.

In this work, we use the conceptualization of gender as a socially situated and constituted construct, with caveats. First, we recognize that it can be difficult to untangle sex and gender [55, 82], similar to how it can be difficult to untangle nature and nurture. Second, pronoun usage is unreliable: "woman" may refer to gender and "female" may refer to sex, but in practice there is little standardization. Third, researchers tend to collect gender and sex as a single variable, without distinguishing between the two [29, 55]. Given that we use self-reported demographics in HCI research, we will assume gender, i.e., the participant's social identity, unless stated otherwise. A further issue, however, is that researchers may not provide a range of gender/sex options when collecting demographics [2]. This is something that we will grapple with when discussing results based on participant gender. However, robots do not have a sex, and their gender identity, if any, is one-sided: up to users, roboticists, and researchers to assign [134]. Indeed, this will be a central issue in our attempts to understand how gender has been framed in the literature.

### 2.2 Gender/ing in and of Socially Embodied Agents and Their Study

Gender/ing of technological artefacts is not a new concept. Decades ago, Harding [47] introduced the notion of gender as a symbolic quality of concepts and abstractions; for instance, gendered metaphors and the gendering of concepts such as objectivity and subjectivity. This idea was later extended to objects, especially technological objects [12, 99]. Objects can be consciously or unconsciously gendered based on their properties or become gendered through use. For instance, a computer used for clerical work may be gendered feminine because of symbolic and societal associations of women with such work. Within HCI, seminal work by Nass, Reeves, Moon, and colleagues has revealed that people tend to unknowingly attribute gender to computer voice when gender cues in pitch and timbre are present—and then react to their ascriptions in gender stereotyped ways [93, 112]. Later work confirmed many of the same effects in scenarios with "bodied" agents like humanoid robots [107, 108]. Following this, a body of work emerged in which robot gender/ing was directly considered, especially in terms of gender stereotyping in robotic design and deployment: femininized robots in home scenarios [20]; perceptions of agency (masculine) and communalness/warmth (feminine) in facial cues [40], waist-to-hip ratio [13]; hats [69]; uncanny reactions [102]; acceptance of help when coded feminine [77]; men donating more money to a robot coded feminine [132]; and occupation assignment [142]. Most recently, the growing availability of conversational agents and voice assistants has led to critiques of the standard "female by default" vocal embodiments that reflect gendered ideas of women as subordinate [54, 129, 145]. In general, this work has confirmed stereotyped gender/ing, with some exceptions (cf. [111]). Men compared to women have been found to prefer robots more, with little known for those of other genders [78, 92, 113, 126]. Also less explored are cultural differences in gender/ing. For instance, Robertson [116] reported on the case of Wakamaru, a robot with a form factor based on traditional men's clothing in Japan that resembles Western petticoats worn by women; subsequently, Japanese people tend to gender Wakamaru masculine, while Westerners tend to gender it feminine.[2] In short, people attribute gender based on the design and/or context of use of robots and other computer-based artificial agents, and often in unconscious, stereotyped ways.

Virtually all of this work has focused on participants as users or potential users. However, engineers and researchers are surely not exempt [116, 134]. A trickle-down effect can occur, where gender is taken for granted based on societal models. In most cultures, "man/male" is the default, assumed when no other cues exist, and sometimes even when they do [6, 48, 127]. This "default setting" then extends to choices in the design and study of socially embodied artificial agents like robots. For instance, Robertson [116] revealed the prevalence of gender stereotypes in Japanese robots, which were gendered in ways that "effectively [reproduced] a sexist division of gendered labor among humans and humanoids alike" [116]. Education, exposure, and community homogeneity also have consequences for approaches to gender [128, 138]. When working with people of diverse gender identity, one is positioned to see gender from new perspectives. When not, one may resort to limited models of gender, without realizing other options exist. More research is needed to map out how this has and continues to play out, especially its extent. Importantly, what little work exists has so far focused on the creator side: engineers and roboticists. After all, it is not difficult to detect when the robot itself is evidence.

As practitioners, we can take on at least three orientations to gender/ing: we can remain unwitting, we can address it explicitly, or we can choose to avoid it. This last point, a "gender-evasive" approach, means consciously choosing to avoid the matter of gender/ing rather than engage with its complexities. But this runs the risk of reinforcing limited and harmful representations of gender that may cascade back to people [136]. While the relationship between sexist portrayals and attitudes, let alone behaviour, remains contentious, some compelling arguments for robots [136] and empirical findings for video game characters [7, 140] exist. Furthermore, the possibility of enacting or enforcing negative associations based on gender/ing needs to be considered against the availability of the agent. For example, Fessler [44] conducted a technical test on how the predominant range of "feminine" voice assistants responded to sexual harassment in 2017, which was later replicated by Chin and Robison [26] in 2020. In the first run, responses ranged from "flirtatious" to evasive: conscious and unconscious but no less sexist scripts that cast the feminine as sexualized and subservient. The follow-up test indicated corrective redevelopment, with responses ranging from dehumanizing the agent ("I'm code"), ignoring the prompt, and declining to respond. Still, while confronting sexism may disrupt attitudes [85], ignoring or declining to respond may not. Nevertheless, this work shows that confronting gender/ing is a step in the right direction.

Avoiding gender in pursuit of a *tabula rasa* or stateless, gender-neutral humanoid robot may also not be possible. For the same reasons others have argued that AI algorithms may always be biased, i.e., fairness may be unattainable [71], robots are *created by people* who live in a gendered world, and so they may always be gendered [134]. Yet, we also live in a world where gender is socially constructed and robots are made: we thus have the power to consciously construct robot bodies [116]. Over a decade ago, Rode [117] raised the possibility of other valid options brought in from critical gender studies, including non-binary genders, gender fluidity, and third genders. In fact, we can view robots as an opportunity to explore new gender forms and expressions. This leaves room for "mechanical genders" [134] that may emerge as robots diverge from human models. As Schiebinger proposed [127], we can promote a "virtuous circle" towards the prosocial goal of equality and recognition of gender diversity. We "… have the opportunity to intervene in this cultural cycle by creating hardware that promotes social equality [..] helping users rethink gender norms and eventually reconfiguring gender norms" [127].

As yet, we know very little about how researchers have approached gender/ing. Gender/ing may take different forms within research, reflecting the larger discourse outside of HCI spaces or pointing to disciplinary alternatives. In this sense, obtaining a better understanding of the state of affairs through the artefacts produced by researchers that are emblematic of their approach/es to gender/ing— namely publications—is a crucial first step. Such a self-critical positioning may be classified as reflexivity [117]. Reflexivity typically involves critical and theory-driven engagement with researchers as social actors and research itself as a social phenomenon [15, 88, 117]. Since publications are largely text-based, we can use the written word as data. Pronoun use and referents based on gender signifiers (e.g., feminine) are clear textual indicators of gendering. Analysis of such patterns in the language used by researchers in publications can thus act as an unobtrusive, objective measure of how gender/ing has been approached.

In short, gender/ing is an important factor to consider in our research. Yet, despite a growing body of work on the participant or user experience, as well as critiques of roboticists' approaches to gender/ing of robots, there is little known about the researcher's side— our side. We argue that it is time to confront this issue by taking a self-reflective stance: unmask how gendering has affected our practice. We take a first step in this paper, using Pepper as a guiding example.

## 3   METHODS

We conducted a systematic review [72, 131] to explore the role of gender and gendering in the case of Pepper from both "our" (researchers, designers, engineers, roboticists, practitioners, etc.) and users' perspectives. We used an adapted [3] version of the PRISMA checklist [81] to structure our process and reporting. Our study selection flow chart can be found in Appendix 1 in the Supplementary

---

[2] Robertson goes on to describe Wakamaru's shifting gender attribution by its makers, a case in reconstituting gender ascriptions within engineering praxis.

[3] The PRISMA checklist is based on norms in medicine, so not all parts are relevant to a CHI paper. For instance, structured abstracts.

Materials. We used a meta-synthesis approach [66], focusing on integrating paper metadata, descriptive statistics, and qualitative findings. We supplemented this with directed content analysis [52, 76] of how gender was treated in-text by the authors as well as in quotes from participants, focusing on pronoun usage and descriptions of gender/ing. We did not use or register a protocol. Our high-level research question was: *How has gender/ing been approached by researchers in research on Pepper?* To answer it, we focused on several sub-questions.

*RQ1. What gender has Pepper been assigned by researchers (if any)?* Mapping out how researchers have gendered Pepper or avoided gendering Pepper will reveal what gender model(s) have been used as well as how Pepper's gender, or lack thereof, has been ascribed by researchers.

*RQ2. What gender has Pepper been assigned by participants (if any)?* Mapping out how participants' gendering of Pepper can provide insight on the extent to which participants and researchers gender the same or differently. It can also answer the basic question of what gender, if any, is most appropriate for Pepper in terms of user experience.

*RQ3. Has there been a manipulation check on gender?* Use of a manipulation check is an indication of researcher's awareness of gender. It also provides more robust findings on participant gendering of Pepper.

*RQ4. What reasons (if any) are given by researchers for Pepper's gender?* Researchers who are oriented towards addressing or explicitly avoiding gender may give their reasons for doing so. This would provide a deep understanding of researcher approaches to gender and how they relate to larger models of gender outside of HCI spaces.

*RQ5. What findings are there based on participant gender?* Researchers' approaches to gender can be revealed through choices in research design, methods, measures, and data analysis based on participant gender. Additionally, what participant genders were accounted for—and which ones were not—can reveal researcher views of gender that would likely relate (in the trickle-down fashion) to their views on Pepper's gender/ing.

*RQ6. What findings are there based on participant perception of Pepper's gender?* As with RQ5, reporting of participant's gendering of Pepper— ideally, how this data was collected—can reveal how researcher's themselves approached gender. For instance, researchers may provide a questionnaire with certain response options and not others.

*RQ7. Is there a relationship between participant and researcher gendering?* Researchers' approaches to gender/ing may influence participants' orientations to Pepper's gender/ing. Great similarities may suggest influence or congruence, while divergence may indicate a lack of influence and greater level of gender consciousness.

Before outlining our methods in more detail, we will first introduce Pepper, the impetus for our study.

### 3.1 Case Study: Pepper the Humanoid Robot

Pepper (Figure 1) is the spiritual successor of the Nao robot, a widely available humanoid robot that has enjoyed a rich life in research and child education [8]. Compared to Nao, Pepper is taller, sports a tablet on its chest, and is described as "socially smarter," i.e., has more sophisticated realtime processing of user-derived data. Unlike its predecessor, Pepper can recognize people and emotions by interpreting faces[4]. Like Nao, Pepper has been taken up as a tool for research, as well as used outside of the academy, especially as a storefront greeter. As discussed, people seem to gender Pepper in widely different ways. For example, Pepper has been gendered as a "young girl" when placed in the role of caretaker within societies where caretaking work is othered and considered "less than," and thus the domain of "anything other than white adult males" [86]. This and other divergent ascriptions of its gender inspired us to empirically explore how gendering plays out in robotics research[5] with Pepper as a case study. Additionally, neither of us has conducted research on Pepper, allowing us to take on a more objective view than we would be able to with robots familiar to us.

### 3.2 Eligibility Criteria

Any human subjects research on Pepper was included. In our first search, we focused on the major engineering and/or computer science publication venues that cover human-robot interaction (CHI, HRI, HAI, RO-MAN, etc.) using IEEE Xplore and ACM Digital Library. We then expanded our search to include venues from any field by using Scopus and Web of Science. We excluded work that did not include human subjects studies, that were proposals of studies, and that were gray literature. Distinctions were made to maximize the papers that would include both researcher and participant perspectives. We limited results to work published within and after 2015, the year that Pepper was first available.

### 3.3 Search Queries and Study Selection

We began with a search of the core databases (ACM Digital Library and IEEE Xplore) on June 4, 2020. We used the following search terms: "Pepper" and "robot" and ("sex" or "gender"). We removed duplicates and papers that did not meet the eligibility criteria. We then read the full text of each paper and removed all that did not explicitly use Pepper (i.e., Pepper as an example, cited for other work, etc.). We then used general databases (Scopus and Web of Science) to search for other relevant work. We followed the same procedure with the resulting papers as before. The databases, queries, and total results at each stage are presented in Appendix 1. On June 22, 2020, we ended with a total of 75 papers.

### 3.4 Data Analysis

One author extracted all data relevant to the research questions. This included: participant demographics (mean age, age ranges, gender), researcher's attribution of gender to Pepper, researchers' reasons for this attribution, participants' attribution of gender to Pepper, study findings related to participant gender and Pepper's perceived gender, and the presence or absence of a gender manipulation check for Pepper. Categories for assigning gender emerged from the use of pronouns in the text, and thus the range of options increased over the course of data extraction.

---

[4] The extent to which Pepper can recognize emotions through facial expressions has been debated [116].

[5] For the sake of transparency and author self-disclosure [128], one of us is in HCI and the other studies robots from cultural and literary perspectives.

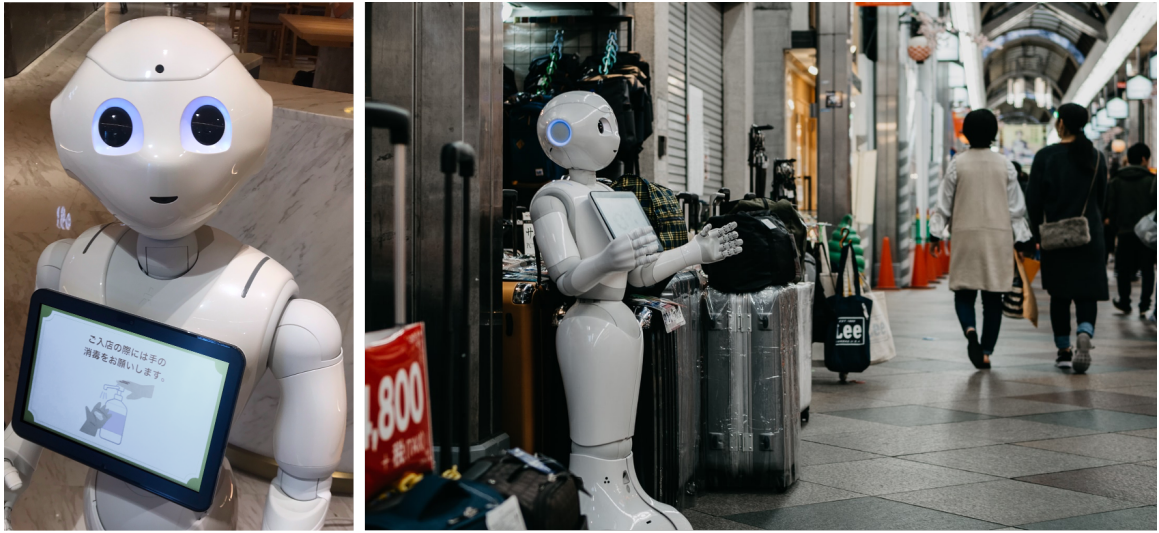

**Figure 1: Pepper close-up (left) and at a storefront in Japan (right). Left photo by the first author and right photo by Lukas (@hauntedeyes), used under the free use license provided by Unsplash (https://unsplash.com).**

For the meta-synthesis, one author generated descriptive statistics for the quantitative variables: means, standard deviations, medians, and interquartile ranges. Quantification of the qualitative data related to researchers' and participants' gendering of Pepper as well as presence or absence of a gender manipulation check was generated in the form of frequency counts and percentages to enable comparisons between the possible options per variable. Both authors then conducted three thematic analyses on researcher's reasons for their attribution of Pepper's gender as well as study findings (divided by participant's gender and Pepper's perceived gender). One generated a set of initial themes based on the research questions as per a typical directed thematic analysis process [17]. Then, both separately coded all of the three sets of qualitative data. Cohen's Kappa was calculated as a measure of inter-rater reliability; the value for each of the included themes in all three sets of data was .79 or above. Where possible, t-tests were used to compare results.

## 4  FINDINGS

We now present the results from our analyses of the surveyed papers. The full list of papers and their metadata is available in Appendix 2 in the Supplementary Materials.

### 4.1  Author and Participant Demographics

In 75 papers, there were 322 authors, 286 of which (89%) were unique, i.e., had only one paper in the set of surveyed papers. Across all studies, there were a total of 3993 participants. The average age of participants was 31.7 (STDEV = 14.4, MED = 27.9, IQR = 49.2), with ages between 4 to 94 (a span of 90 years). 17 papers (22.7%) did not report on age. 1810 men (45.3%) and 1599 women (40%) participated, with 11 undeclared or attributed as "other" (0.3%); transgender, gender fluid, agender, and other genders were not reported. 14 papers (18.7%) did not report participant gender.

### 4.2  Researchers' Gendering of Pepper (RQ1)

In most papers (47 or 63%), researchers attributed object gender to Pepper, using the pronoun "it." In 26 papers (35%), no gender attribution was found; researchers used Pepper's name or "the robot" instead of pronouns (e.g., [61, 87, 109]), or used multiple robots and so avoided pronouns (e.g., [110]). In three cases, Pepper was gendered feminine, in two cases masculine, and in one case neutral. In some cases, assignment was enforced via gender cues or by modifying Pepper; for instance, the authors of [38] renamed Pepper as "Tina" so as to "contextualize its gender as female" (p. 244). In 4 cases (5%), researchers switched their use of pronouns: two cases of feminine pronouns and object pronouns [37, 120], and two cases of masculine pronouns and object pronouns [21, 30]. Researchers did not consider Pepper's gender in any of these cases. Indeed, the authors of only 12 papers (16%) recognized and openly discussed Pepper's gender. Of these, the majority (8 out of 12) referred to Pepper as an "it." Of the rest, one attributed a neutral gender to Pepper and the others avoided attribution. Non-textual cues to Pepper's gender could be present, but this was hard to extract and confirm, especially in the face of conflicting evidence. For instance, the authors of [80] used objective pronouns for Pepper, but also compared Pepper to a virtual woman character and a real woman. This suggests that they viewed Pepper as feminine or fluid, leaving interpretation of gender to comparison. In another case [130], Pepper was called a "semi-humanoid," which could imply that the authors believed that certain characteristics such as gender do not apply to Pepper.

### 4.3 Participants' Gendering of Pepper (RQ2)

In most papers (52 or 69%), participant attribution of Pepper's gender was not available, i.e., direct quotes were not reported in the paper. Of the rest (23 papers), 15 papers (65%) included object gendering, 13 papers (57%) included masculine gendering, 8 papers (35%) included feminine gendering, and 8 (35%) included neutral gendering (e.g., use of "they" and "them" as a singular pronoun). For example, in one paper [101]: "Pepper's response was suddenly delayed, so I thought he could not hear well" (p. 197). Just over half (57%) of papers included participants gendering Pepper differently, e.g., one participant used object pronouns while another used feminine pronouns. For example, in one case [104], most participants used object pronouns, but one used a neutral pronoun: "I don't like to be disrupted by them" (p. 8): perception of Pepper's gender depended on the participant. Notably, children as young as four years old attributed gender to Pepper. For instance, in one paper [156], children used the masculine referents "sir" and "he." Children also asked if Pepper had a wife—an implicit masculine gender cue, given the above. It is possible that participants were influenced by researchers' use of pronouns for Pepper in these studies, which we cannot confirm using our methods.

### 4.4 Gender Manipulation Check (RQ3)

Only 6 (8%) papers reported on a manipulation check for Pepper's perceived gender. Put another way, 92% of papers did not report on (and we assume did not conduct) a manipulation check. Of those that did, most studies conducted a typical check: before the main study, with participants. One study [18] involved a post-hoc check with only those participants who agreed with the researchers on their own gendering of Pepper in various conditions; this of course means that their results are not representative. Another study [102] involved researchers tasked with independently assigning gender to multiple robots, including Pepper, to build a consensus about each robot's perceived gender. When they could not agree, the researchers assigned a "neutral" gender label, including to Pepper. We point out that this use of "neutral" does not match most definitions of the word; more fitting labels could be "undetermined" or "contested" or perhaps "fluid." We will reveal more cases and continue to tease out issues in researchers' reasons for gendering Pepper next.

### 4.5 Researchers' Reasons for Pepper's Gender (RQ4)

The majority, a total of 50 or 67% of papers, did not discuss Pepper's gender. In 12 papers (16%), the authors acknowledged the issue of gender but did not act on it; for example, by citing other work (e.g., [24, 36, 91, 100]) or acknowledging gender as a factor for future work (e.g., [31, 51]). 8 papers (11%) provided some kind of reasoning, but it was insufficient to understand their assignment of gender. For instance, there were claims that the TTS used to produce Pepper's voice is childlike (e.g., [37, 57], with the implication that children do not have gender or are gender-neutral (e.g., [89, 104]). Other reasons varied. For instance, the authors of [67] gave their robots "unfamiliar male Hawaiian names ... to reduce possible gender biases" (p. 590) without explaining how this would reduce, rather than introduce, a male gender bias. In [23], the authors focused on "appearance," excluding gender as an aspect of appearance.

The authors of 7 papers (9%) noted that gender cues existed, but there was a lack of consensus across papers about what gender/s those cues cued. One case [10] interpreted Pepper as gender neutral and then explained participants' gendering of Pepper as male as "probably related to its physical appearance, because participants look at only a picture of the robot, without interacting with it. Therefore we guess that Pepper robot is identified as male because of its body shape" (p. 42). The authors of [121] found that participants gendered Pepper despite impressions from a previous study where participants would not. In contrast, the authors of [89] suggested that their participants' gendering of Pepper as feminine may have been due to its "feminine quality of having a smiling expression" or "round" appearance (p. 217).

Pepper tends to be seen as masculine or feminine, despite authors' ascriptions or efforts to present Pepper as gender neutral. In 6 papers (8%), the authors claimed that Pepper is gender neutral. For example, the authors of [10] stated that "we focused on a robot that has no sex and we do not apply any kind of alterations that lead people to attribute easily a sex to a robot" (p. 42). Similarly, another case [102] relied on the premise that robots are gender neutral unless gender cues are added: "robots without gender cues" such as "facial features, hairstyles or a voice" (p. 218). However, their independent raters did not agree with this perspective, with most *not* classifying Pepper as gender neutral. In one case [121], participants were asked to design a robot through illustrations and storyboards; in reviewing these, the authors decided that participants "did not provide explicitly gendered cues" and decided to make a gender-neutral version. While they do not explain why, they equated a high pitch voice with a gender-neutral voice, relying on that along with Pepper's "boxy, simple body" to enforce a gender-neutral presentation. However, participants did not see Pepper as gender neutral: "in interacting with the robot, participants tended to use a variety of pronouns" (p. 8). The authors of [18], in contrast, explored several voices to pinpoint those that could be red as feminine, masculine, or gender neutral.

A final 2 papers made claims of language or translation issues. In [156], the authors decided to use masculine referents for Pepper after translating to English because "children referred to robot at [sic] 'he' due of the [sic] Polish grammar" (p. 5). The authors of [43] avoided the issue of gender in the Hebrew language, which is connected to verb conjugation, by inputting gender as a variable for Pepper to act on.

Considering the above and 4.2, it seems that different people interpreted Pepper's gender differently. Some of the same reasons were used by different researchers to justify these positions. Pepper's gender and aspects thought to indicate gender (e.g., voice, body shape) appear to rely on the perception of the individual.

### 4.6 Findings Based on Participant Gender (RQ5)

About half (49% or 37) of the surveyed papers reported but did not use participant gender in their designs or analyses (e.g., [3, 35, 59, 73, 74, 84, 98, 123, 139, 147, 152, 155]). A further 20% (15 papers) did not mention participant gender at all (e.g., [49, 53, 90]). In some cases, gender was used in other ways; for instance, in [73], the researchers customized Pepper's speech to refer to participants' gender, while in [9] the authors used Pepper's built-in "gender estimation" data stream. About 16% (12 papers) conducted analyses based on participant gender but found no effects. For instance, one study [10] found that men and women attributed gender to Pepper the same way. Another [137] found no differences on willingness to sign a petition between men and women, whether the petitioner was a human or Pepper. In their exploration of the moral agency of robots, the authors of [62] found no difference based on participant gender.

In contrast, 11% (8 papers) did find effects. One [56] found that when people lied to Pepper, their heart rate was higher, regardless of their gender. For example, when the stress level of a task varied, men tended to be more empathetic towards Pepper in less stressful situations, but less empathetic towards Pepper in stressful situations, with the opposite being true for women [106]. One study [67] found that boys preferred the "introverted" robot and more girls preferred the "extraverted" robot. In [122], while watching a movie together, people paid attention to Pepper the same, but women expressed more joy and smiled more when Pepper reacted oddly. While the authors concluded that "men did not let themselves be distracted from robot inconsistency" (p. 4), implying that woman are easily distracted and men have greater self-control, we point out that other interpretations are possible, such as women having a better sense of humour.

Several papers reported on results despite methodological and analysis weaknesses. In particular, relying on non-statistically significant results (e.g., [106, 149]) and running analyses despite uneven groups (e.g., [14, 33, 101]). For example, despite finding no effect based on participant gender, the authors of one study [106] concluded that "an engineer could design less human-like responses for women operators and more human-like responses for male operators based on the task" (p. 5). We decided not to report on the results of studies with such limitations. Indeed, we caution against such practices: we must learn to accept null results and/or strengthen our research, e.g., by recruiting more participants.

When gender was considered, almost all assumed a binary model (e.g., [11, 70, 79, 83]). Some used a male-as-default/male-as-neutral model, e.g., the authors of [22] referred to the generic participant as "he/him." Only 4 (5%) adopted an inclusive model. Some (e.g., [146, 153]) used "they/their" pronouns to refer to participants, while others (e.g., [18, 124, 148, 149]) provided non-binary gender options for participants.

Given the noted issues in collection and analysis, it is difficult to draw general conclusions about whether and when participant gender matters.

## 4.7 Findings Based on Pepper's Perceived Gender (RQ6)

The vast majority (about 87% or 65) of papers did not look at how perceptions of Pepper's gender influenced results. 5 papers (7%) explored Pepper's gender from a binary perspective, and 3 papers (4%) explored inclusive genders. 4 of these (5%) found an effect. One [119] found that only 15 of 50 participants in Pepper's "gender neutral" condition agreed with that gender assignment. Other results from their study aligned with gender stereotypes, e.g., participants preferred a robot instead of a woman for package delivery and a robot instead of a man in a receptionist position. Similarly, one [10] found that male participants reported that Pepper would be best for what they argued are stereotypically feminine tasks, i.e., reading fairy tales, despite participants assigning masculine or neutral genders to Pepper. This could mean that such tasks are not necessarily feminine, or that Pepper's gender was more complex or even fluid, depending on the task. The authors of [102] found that feelings of uncanniness about Pepper only occurred when Pepper was "gendered" either masculine or feminine. The authors of [38] found that feminizing the robots they used led to them being better accepted by the older adult and elderly participants. In [18], researchers found no effects based on Pepper's perceived gender; rather, other aspects of performance affected results.

Methodological limitations and lack of detail obscured understanding of some results. For example, the authors of [141] found that feminine robots received more sexual and negative comments than male robots. However, they did not state which robots they had categorized as feminine. They also could not collect participant gender because they used comments posted online by unknown, semi-anonymous parties as their data source. In another case, the authors of [143] wrote that "this study has not taken any measures to ensure that the effects are actually caused by the different degrees of anthropomorphism [...] this leaves open the possibility that other properties of the robots are the true causes of the observed effects" (p. 300). Another [105] analyzed Pepper's perceived gender but did not account for participant gender. The authors of [154] read gender into children's descriptions of Pepper: "creepy is a gendered term; people perceive men to be creepier than women, particularly as a sexual threat" (p. 3). However, this seems to refer to adults' mental models of the world, so to apply it to children is a leap, and the authors do not provide a source for this view.

Some who did not explore Pepper's gender discussed it in other ways. For instance, the authors of [151] included gender in their proposed model of a robot persona; it is not clear what results, if any, this decision was based on. The authors of [156] noted, without detail, "[points] of interest on Pepper's body ... [the] issue of embodiment needs further investigation" (p. 5). Others (e.g., [75, 89, 143, 144]) referred to Pepper's anthropomorphism or social influence but without explicitly mentioning gender. Still others cited literature on robot gendering (e.g., [36, 91, 94, 125]), but did not consider it as a factor in their own work.

Given the lack of gender manipulation checks in most studies (see 4.4), it is possible that participant perception of gender was a latent variable that could have been influenced by researcher decisions about Pepper's presentation (e.g., voice, body modifications) as well as its role or the task/activity.

## 4.8 Relationship Between Participant and Researcher Gendering (RQ7)

Of the 23 papers in which participants ascribed gender to Pepper, 8 (35%) papers demonstrated that researcher ascriptions matched those of participants exactly, i.e., both used the same pronouns or gender signifiers for Pepper. In the other 15, the researchers' gendering was among those used by participants in 5 (33.3%) cases. Of the 11 cases where researchers addressed gender/ing directly and provided data about participant gendering, 8 (73%) showed diverse participant ascriptions of gender, with the other three matching the researchers' ascriptions exactly. Of the 12 cases where researchers did not explicitly address gender and data about participant gendering was available, 7 (58%) matched exactly, 3 (25%) matched partially, and two did not match. A t-test comparing number of matches between researcher and participant gendering based on researcher recognition of gender/ing showed a significant difference, $t(10) = -2.07$, $p = .03$, $d = 1.33$. These results suggest that there was some influence of researcher gendering of Pepper on participants, assuming participants were faithfully recorded and did not in fact influence the researchers themselves. Additionally, researcher recognition of gender/ing appears to have played a role, with significantly more exact matches between researcher and participant ascriptions of gender when the researchers did not directly consider gender/ing.

## 5 DISCUSSION

Our results overwhelmingly demonstrate that Pepper's perceived gender is not a simple fact, nor without consequences. We have summarized our main findings in six challenges underlying this state of affairs. We then offer a practical framework to guide consideration of gender/ing in research praxis. We also provide a critical self-reflection on terminology choice, disciplinarity and readership, and walking the line between science and humanities approaches.

### 5.1 Invisibility

Gender/ing was often invisible, having been taken for granted or assumed [115]. This means that gender was not used as a factor of design or analysis, or critically interrogated. Additionally, whatever influence (or lack thereof) gender/ing had is largely unknown. Gender/ing can become visible through participant data (such as through the use of multiple pronouns), through post-hoc reflective work on the part of researchers, through follow-up work (such as running gender manipulation checks), or through outside critique (e.g., the present work). Pronoun switching can indicate gender invisibility on the part of the researchers. A lack of gender content can indicate null results, but we would argue that these should be reported as a part of establishing knowledge about the influence (or lack thereof) of gender/ing.

### 5.2 Variability

One effect of gender invisibility was variability in Pepper's perceived gender. In many cases, we did not have access to how Pepper was or was not gendered, by researchers or participants. In the rest, perceptions of Pepper's gender varied from objective to masculine, to feminine, to neutral, to fluid (depending on the task and context), to multiple attributions (whether or not participants and researchers were aware of it). In contrast, researchers' models of participant gender tended to be binary. This means that researchers' implicit or explicit models of gender did not quite match what was really going on in the world. This also means that a broader view of gender is necessary. The implications for a broader view of gender extend beyond Pepper and other agents. In particular, the act of recognizing the variability in perceptions of Pepper's gender may inspire researchers to take a more diverse perspective on participant gender. As we found, reporting of participant gender was largely limited to the gender binary, "other," or unstated categories. Very little representation of other possibilities, including transgender, gender fluid, agender, and third genders, existed, despite a growing recognition of genderful folk and, with increasing societal acceptance, comfort in being out [46].

### 5.3 Decentrality

Decentering gender is the opposite of centering gender. By decentering gender, we mean acknowledging but not acting on gender as a nucleus of study. In the surveyed papers, this appeared in several forms. One was the act of citing and discussing other work that addressed gender, without doing the same. Another was recognizing the importance of gender after the main work was done and proposing future work. Still another was a kind of *agender anthropomorphism*: considering the human-likeness of Pepper without discussing gender (i.e., what "points of interest on Pepper's body" were there?). A more subtle form was through researchers' insufficient justifications of their limited approach to gender. More subtle yet, relegating these to footnotes. All of these are ways in which gender was moved from the center to the sidelines. In this way, decentering gender contributes to gender invisibility and variability—moving it away from view, providing it no structure. The reasons behind decentering are unclear, and could span a range of topics, e.g., hesitation due to lack of knowledge or training about gender, prioritization of research matters, lack of awareness, etc. A next step for future work is to unearth these reasons, such as through interviews, reflexivity activities, and so on.

### 5.4 Neutrality

A common theme in ascribing Pepper's gender was "neutrality." Yet, most of the literature showed a failure to find agreement among participants when it came to attributing a "neutral" gender to Pepper, even when given the option and even when tested against masculine and feminine alternatives. This and Pepper's Japanese origins point to the idea of mukokuseki, or "statelessness," a term coined by Koichi Iwabuchi [60] to describe the "cultural vagueness" of Japanese exports. In theory, mukokuseki aims to break borders and enable interpretability. For example, trademark "anime-isms," like big eyes and colourful hair, are meant to be digested by non-Japanese audiences according to their own culture's notions about appearance. This has led to some long-standing debates, such as the "Are anime characters white?" conversations that continue to percolate through social media and online forums. Still, mukokuseki design has helped anime, manga, video games, and electronics to flourish overseas. Pepper also appears to be stateless in appearance. Although primarily used in Japan, Pepper is cute, appealing, and ambiguous in a way that makes it an ideal export. Yet, mukokuseki products cannot control their reception or interpretation when sent abroad. Even if genderlessness was the design team's intention, as they have claimed, the surveyed findings show that Pepper's design communicates a range of genders. This is not to say that Pepper's design should be re-interpreted or more clearly marked "masculine" or "feminine." Mukokuseki design raises questions about gender and other markers, such as race, age, and nationality, by pushing researchers to observe the inevitable biases that come into play through the non-interrogation of these very aspects. We cannot assume that Pepper's pronouns should be neutral, and this is likely true for other agents.

Seaborn & Frank (2022)

## 5.5 Convergency

Latent factors that converged with gender but were unexamined by the researchers seem to have influenced perceptions. Age was most apparent, both in terms of participant age and how gender was conceptualized and described, generally within the paper or specifically with respect to Pepper. Children as young as 4 years old were found to gender Pepper, and gender was collected about these child participants. Yet, some researchers interpreted Pepper's default voice as both childlike *and* genderless. What this suggests is that age and gender need to be interrogated together. Gender does not disappear as a result of age, but age can mask the issue of gender and catch us unaware. Some work also pointed to a convergence of factors related to language and translation. For instance, explicit decisions were made to use masculine identifiers as gender-neutral referents (e.g., Hawaiian male names, Polish pronouns) without a reason why. Language can shape and change our perceptions of the world. For instance, previous work has shown that people are more likely to gender even gender-neutral objects and beings when their language uses gendered pronouns, i.e., like English, compared to those who use "epicene," or genderless, languages, i.e., Karitiâna [39]. We were lucky to find examples in the surveyed work, made possible by the authors' transparency. We suspect that there are many more latent factors at play, but we are unable to spotlight them due to their very nature as latent. Even so, this presents an opportunity for cross-disciplinary work involving linguists, gerontologists, critical theorists, and others who may have insight into what latent factors converge with gender. Possibilities can then be explicitly studied and confirmed or rejected.

## 5.6 Circumscriptivity

Gender was circumscribed in ways related to the embodiment of Pepper and larger models in society. One of the clearest ways in which this played out was a reliance on dominant models of gender. This involved focusing on binary models, conflating sex and gender, and not considering the nuanced relationship between sex and gender. For instance, voice was an important aspect of Pepper's embodiment. The results showed that voice was consciously used (and even changed) to influence binary attributions of gender to Pepper and gender stereotyped perceptions, attitudes, and reactions to Pepper. Gender being decentered and invisible contribute to how it has been circumscribed. Even when gender was considered explicitly, there was a general exclusion of models and perspectives that deviate from the norm. With some exceptions (e.g., [18, 102, 121]), researchers made decisions about their research and its results in ways that appear to align with their (almost entirely undiscussed) models of gender alone. Finally, we found some evidence that researchers' gendering of Pepper influenced how participants gendered Pepper, especially when researchers did not explicitly address gender. We recognize these as issues of power. As researchers, we are in a position to determine how participants contribute to our research by limiting the options we provide for them. For instance, we can unintentionally prime participants through our own unconscious gendering of the agent expressed in our design of response options on questionnaires. We provide some options to avoid this in the framework presented next.

## 5.7 A Practical Framework

Even while most studies did not treat gender/ing as a key factor, the results indicate that we must. Yet, we may not think to consider gender/ing when not conducting studies on gender/ing. Moreover, those of us in engineering and computer science may not know about or be exposed to the idea of gender/ing. Disciplinary norms have been shown to play a key role in what is recognized, valued, and promoted, and technical fields, like robotics, are no exception [16, 25]. Even if we are on board, we may not have the background to ensure that we have considered it fairly. As a baseline, we advocate for taking a gender neutral stance, especially in language use, so as to remain open and nonprescriptive. As a means of centering gender, i.e., targeting the problem of gender being decentred, we offer this framework for researchers working on socially embodied artificial agents like Pepper. It is built upon the critical insights about gender/ing that we uncovered in this work using Pepper as a guiding example. We first present the centerpiece of the framework: a critical checklist geared towards researchers and research. We then provide strategies and examples that researchers can use to help answer or take action on these self-check questions. We note how each addresses the six challenges we identified above.

### 5.7.1 A Critical Checklist

In Table 1, we provide a list of self-check questions that all researchers can ask themselves in the context of any research project involving socially embodied artificial agents. We have marked the questions that we feel are essential even in cases where gender is not a locus of study. Each question was derived from our results, based on the gaps, inconsistencies, assumptions, and limitations we discovered. Answering these questions or taking action where needed will center gender/ing as well as raise the quality of the research produced. While this checklist may not cover every issue or guarantee a perfect approach, it should go some way in increasing rigour when it comes to gender/ing.

### 5.7.2 State Your Position on Gender/ing

Most of the surveyed work, even ones that considered gender, did not state a position on gender/ing. In this, there are at least two perspectives that need to be addressed: (1) gender as a characteristic and (2) our propensity to gender agents and their embodiment. Gender as a characteristic applies to participants (i.e., as a demographic variable), ourselves, and the agent (i.e., gender cues in its design and context of use). We do not have to keep to conventional models or academic research; gender is a dynamic factor and so established models may not match participant—or our own—experience. If our positionality on gender differs between participants and the agent, we should state so. All models and theory should be cited. Ideally, there should be justification for the choice of models and discussion of alternatives.

*Challenges Targeted:* Invisibility, Variability, Neutrality

*Example:* We use the conceptualization of gender as a socially situated and constituted construct (continued in 2.1). We both identify as cisgender women, and one of us leans towards gender apathy.

**Table 1. Critical self-check questions, listed by focus and related to the challenge/s each target and strategies, where applicable. Questions deemed essential even in cases where gender is not a focal point are also marked.**

| Focus | Questions | Strategy | Challenge/s Targeted | Essential |
|---|---|---|---|---|
| Researcher/s | What model/s of gender have we used, in general and for the agent? | 5.7.2 | Invisibility, Neutrality | · |
|  | Have we gendered the agent under study, in its design, context of use, or descriptions in ethics forms, scripts, or other materials? | 5.7.7 | Invisibility, Neutrality | ○ |
|  | If so, what gender/s? | 5.7.2 | Variability | ○ |
|  | Have we been consistent with our pronoun use and use of other gender signifiers? | 5.7.7 | Variability | ○ |
| Participant Gendering | Have we captured participant gender? | 5.7.3 | Invisibility | ○ |
|  | Have we allowed for a range of responses, regardless of our own positioning? | 5.7.4 | Variability, Neutrality, Circumscriptivity | ○ |
|  | Have we considered asking participants about how they conceptualize gender, in general and/or with respect to agents, without enforcing our own views? | 5.7.4 | Variability, Neutrality, Circumscriptivity | · |
| Agent Gendering | Have we included a way to capture participants' gendering of the agent, if it occurs, without assuming it does? | 5.7.4 | Variability, Neutrality | ○ |
|  | Have we allowed for a range of responses, regardless of our own positioning? | 5.7.4 | Variability, Neutrality, Circumscriptivity | ○ |
| Data Analysis | What gender/s is/are the agent perceived to be by participants, if any? | 5.7.5 | Invisibility, Variability, Neutrality | ○ |
|  | What gender referents have been ascribed to the agent? | 5.7.5 | Invisibility, Variability, Neutrality | · |
|  | Did participant gendering match that of the researcher/s? |  | Circumscriptivity | · |
|  | What models or novel ideas about gender emerged? | 5.7.4 | Variability, Neutrality, Circumscriptivity | · |
|  | How did participants react to the agent, especially in any stereotyped ways? | 5.7.6 | Variability, Circumscriptivity | ○ |
|  | What other factors may have influenced or confounded these results? |  | Convergency | · |
| Reporting | Did we report on gender demographics, including genders not covered? | 5.7.3 | Invisibility, Variability, Circumscriptivity | ○ |
|  | Did we report on all results related to gender, including non-significant results? | 5.7.6 | Invisibility | · |
|  | Did we check our paper and other materials to be published for consistent gendering of participants and the agent under study? | 5.7.7 | Invisibility, Neutrality | ○ |

*Seaborn & Frank (2022)**5.7.3 Document Participant Gender*

Even though effects based on participant gender were found, a portion of the surveyed literature did not collect and/or report on participant gender. This should be a baseline demographic variable in all such research. As our findings show, the gender of participants (and likely ourselves) can have real effects on research outcomes, including gendering.

*Challenges Targeted:* Invisibility

*Example:* Collect and report on participant gender as a part of demographics.

*5.7.4 Use Gender-Expansive Framings*

Regardless of our own positioning or the expected positioning of participants, we should allow for breadth of gender/ing. Partly this is to avoid unintentionally biasing participants. Our findings revealed some evidence of a link between researcher gendering of Pepper and participants' gendering of Pepper. This was pronounced in cases where researchers did not recognize gender/ing, but also seems to have occurred even in cases where gender/ing was recognized. Additionally, participants ascribed a range of genders to Pepper; Pepper was not necessarily seen as neutral or agender, even among participants in a single study. It is possible participants will ascribe gender options that we cannot imagine and account for in advance. Finally, participants may have a gender identity that does not match our positioning; the limited results for non-binary genders is telling. To counter these issues, we can adopt a *gender-expansive* framing. In practice, we can allow participants to tell us how they are gendering themselves and a given agent, as freely as possible.

*Challenges Targeted:* Variability, Neutrality, Circumscriptivity

*Example:* We recommend an item and response set similar to the following, which is based on theories of gender, gender critical approaches to humanoid robots, and how gender has been formulated in the Pepper literature so far:

How would you gender the agent, if at all?
- Genderless or neutral
- Feminine
- Masculine
- Somewhere in between feminine and masculine, having characteristics of both
- Fluid, changed over time
- Mechanical
- Another gender: [fill in here]

We do not propose this as a final set. We offer this as a format and starter set to provoke discussion. A standard approach to capturing perceptions of agent gender will be useful for consensus building and future survey work, i.e., standardizing practice for optimal knowledge construction. We suggest developing a standard for future research. Additionally, these categories may be too limited to fully understand participants' models of gender and/or approach to gendering. Open-ended questions could be used to gather information on how participants conceptualize gender.

*5.7.5 Confirm Perceptions of Gender and Gendering*

We should confirm whether and how gendering occurs. One standard approach is a manipulation check [50]. This can be done in a pilot study or pre-study, especially if a gender is a key factor or when theory or previous research suggests that perceptions of gender will have an effect. But a simple post-study (to avoid priming effects or stereotype threat) item recording perceptions of gender may be fine in most cases. This approach is somewhat limited by its subjectivity, however. A more objective approach could be a content analysis of pronoun usage and gender signifiers.

*Challenges Targeted:* Invisibility, Variability, Neutrality

*Example:* The content analysis in the present paper.

*5.7.6 Report All Findings on Gender/ing*

Our findings suggest that gender/ing should be included as a basic variable of analysis. Yet, in most cases, gender/ing was not reported. It is possible that researchers conducted some kind of analysis but did not report it. This may have been due to fear of publication bias against non-significant results, a long-standing problem [133]. In that case, we suggest that authors include a statement that gender/ing analyses were done, with no significant results found.

*Challenges Targeted:* Invisibility

*Example:* We conducted a manipulation check of Pepper's perceived gender, but all participants indicated that Pepper was agender or gender neutral.

*5.7.7 Ensure Consistency in Gendering*

Findings pointed to some incongruency in how researchers gendered Pepper, even within a single study. A consistency check in terms of pronoun use and gender signifiers (in text, but potentially also in images, audio clips, video, or other media, including the agent's speech) may be useful. For greater objectivity, a third party can do the check.

*Challenges Targeted:* Variability, Neutrality

*Example:* Ask a colleague to review the paper for inconsistencies in gendering. Be sure that they know your position.

## 5.8 Critical Self-Reflection

We would be remiss if we did not critically reflect on our own positioning and process. We were two people coming together from very different backgrounds to tackle an interdisciplinary topic in a critical way. The challenges we encountered ran the gamut from terminology to disciplinary conventions to readership. Terminology choice was tricky. We realized that we had to decide about using or not using pronouns for Pepper: Should we avoid explicitly gendering Pepper by using "Pepper" repeatedly? Would the awkwardness of repetition be better than taking a position on Pepper's gender? For ease of reading, we settled on using grammatical object referents, e.g., "its." To the best of our knowledge, this is the most neutral referent in the English language for agents that have not achieved personhood by most measures. Our challenges with terminology were not limited to Pepper's pronouns, however. Reviewers pointed out a clash arising from disciplinary conventions between engineering and the humanities, notably regarding the words "latency" and "decentralization." Our use of these words may be at home in a gender theory paper. But in an HCI paper, even a critical computing one, they may cause confusion due to their sameness/similarity with computer science terms. We realized that we could not simply import terms from other disciplines and let them stand, at least without comment. This also relates to consideration of our prospective readers: What assumptions could we make about what they know, how could we fill in gaps, and what extra-disciplinary framings would be accepted? For example, the topic of cultural tropes about Japan may not be familiar or seen as relevant to STEM readers who may or may not consider how gender dynamics in that country informed Pepper's design. The surveyed papers did not address this, making it difficult for us to use our empirical data, but also giving us pause when drawing conclusions in ways that would be acceptable for critical theorists, anthropologists, and others in critical cultural studies that may not be acceptable in STEM fields. Ultimately, we decided that skating the boundaries between the sciences and the humanities has considerable merit for starting dialogue and inspiring a different viewpoint; we pushed forward despite these risks. We await to see what discourse emerges.

## 5.9 Limitations and Future Work

This work was affected by the COVID-19 pandemic [42]. One of us was repatriated and had our funding nullified, while both of us grappled with the suspension of research as usual, e.g., ethics board closures, bans on human subjects work. We were not able to source all potentially relevant research materials and researcher demographics. In particular, we could not extract author gender. We also did not have access to how gender was handled in research materials for participants. We recognize that additional efforts are needed to determine whether researcher gendering was purposive or unintentional. Gendering or its absence in the published papers may reflect an effort to write with scientific objectivity, possibly influenced by editorial assistance. Author surveys and/or interviews could provide insight on these matters as well as how gender is/was framed, why it was considered or not, etc. We also acknowledge that focusing on one agent (Pepper) limits the generalizability of our findings. Future work should critically review other cases.

## 6 CONCLUSIONS

Our critical exploration of the humanoid robot Pepper has revealed how gender/ing can be an important, influential, and overlooked factor of research practice. This work represents a first step towards a more self-reflective stance on the part of us researchers. As shown through the empirically *non-gender-neutral* case of Pepper, a gender-evasive or gender unconscious approach has limited our research. We have outlined six reasons underlying this state of affairs: invisibility, variability, decentrality, neutrality, convergency, and circumscriptivity. We have provided a practical framework for more critically and deeply integrating gender within research practice. There is great risk of missing out on what is really going on in human-agent experiences as well as perpetuating gender stereotypes, sexist tropes, and unequal representation. We have provided a way forward through clear and reasonable changes to typical practice in terms of theory engagement and methodology. We commit to taking this on in our future work—and we hope you will join us.

## ACKNOWLEDGMENTS

Our gratitude to M.R. Ganda and Peter Pennefather, who reviewed early drafts of his manuscript. We also thank the anonymous reviewers for their careful and insightful feedback.